\documentclass[sn-mathphys,Numbered]{sn-jnl}

\usepackage{graphicx}%
\usepackage{multirow}%
\usepackage{amsmath,amssymb,amsfonts}%
\usepackage{amsthm}%
\usepackage{mathrsfs}%
\usepackage[title]{appendix}%
\usepackage{xcolor}%
\usepackage{textcomp}%
\usepackage{manyfoot}%
\usepackage{booktabs}%
\usepackage{algorithm}%
\usepackage{algorithmicx}%
\usepackage{algpseudocode}%
\usepackage{listings}%
\usepackage{mathtools}
\newcommand\scalemath[2]{\scalebox{#1}{\mbox{\ensuremath{\displaystyle #2}}}}

\theoremstyle{thmstyleone}%
\theoremstyle{thmstyletwo}%
\usepackage{hyperref}
\theoremstyle{thmstylethree}%

\begin{document}

\title[Article Title]{Optimal Kinematic Design of a Robotic Lizard using Four-Bar and Five-Bar Mechanisms }


\author*[1]{\fnm{Rajashekhar} \sur{V S}}\email{vsrajashekhar@gmail.com}

\author[2]{\fnm{Debasish} \sur{Ghose}}\email{dghose@iisc.ac.in}

\author[3]{\fnm{Arockia Selvakumar} \sur{Arockia Doss}}\email{arockia.selvakumar@vit.ac.in}

\affil*[1,2]{\orgdiv{Aerospace Engineering Department}, \orgname{Indian Institute of Science}, \orgaddress{\street{CV Raman Road}, \city{Bangalore}, \postcode{560012}, \state{Karnataka}, \country{India}}}


\affil[3]{\orgdiv{School of Mechanical Engineering}, \orgname{Vellore Institute of Technology}, \orgaddress{\street{Kelambakkam - Vandalur road}, \city{Chennai}, \postcode{600127}, \state{Tamil Nadu}, \country{India}}}


\abstract{Designing a mechanism to mimic the motion of a common house gecko is the objective of this work. The body of the robot is designed using four five-bar mechanisms (2-RRRRR and 2-RRPRR) and the leg is designed using four four-bar mechanisms. The 2-RRRRR five-bar mechanisms form the head and tail of the robotic lizard. The 2-RRPRR five-bar mechanisms form the left and right sides of the body in the robotic lizard. The four five-bar mechanisms are actuated by only four rotary actuators. Of these, two actuators control the head movements and the other two control the tail movements. The RRPRR five-bar mechanism is controlled by one actuator from the head five-bar mechanism and the other by the tail five-bar mechanism. A tension spring connects each active link to a link in the four bar mechanism. When the robot is actuated, the head, tail and the body moves, and simultaneously each leg moves accordingly. This kind of actuation where the motion transfer occurs from body of the robot to the leg is the novelty in our design. The dimensional synthesis of the robotic lizard is done and presented. Then the forward and inverse kinematics of the mechanism, and configuration space singularities identification for the robot are presented. The gait exhibited by the gecko is studied and then simulated. A computer aided design of the robotic lizard is created and a prototype is made by 3D printing the parts. The prototype is controlled using Arduino UNO as a micro-controller. The experimental results are finally presented based on the gait analysis that was done earlier. The forward walking, and turning motion are done and snapshots are presented.}


\maketitle

\section{Introduction}
Bio-inspired robots that can mimic the walking of reptiles can be used for search and exploration, toxic detection missions, surveillance activities and carry low payloads. Among the repliles, lizards are versatile in the sense that they can walk in cluttered environments, climb walls and ceilings. Although lizards are not amphibious, Basilisk lizards have the capacity to walk on the surface of water \cite{glasheen1996hydrodynamic}. Inspired by these lizards, robots have been made that can run on water \cite{floyd2006novel}. The dynamic modeling of quadruped inspired by the lizards have been done in \cite{park2008dynamic,park2009dynamic}. Taking inspirations from the biological lizards, one can build robotic lizards that can be controlled as per our needs during the missions.

The robotic lizards are built using mechanisms that are mostly closed-chain in nature. In the works of \cite{xu2013bio}, Watt-I planar linkages were used to mimic water running lizards. Kinematic analysis and CPG based fuzzy controller design were done along with experimental validation. The one degree of freedom legged mechanisms are commonly used as they consume less power and obtain the required trajectory. It is also easy to control a one degree of freedom mechanism. The most commonly used is the four-bar mechanism \cite{li2012jumping}. Six-bar mechanisms \cite{zhang2019mechanism} and eight-bar mechanisms \cite{haldane2016robotic} are also used as legs. In a gecko robot that climbs vertical surfaces, two degrees of freedom legs are used \cite{kim2008smooth}. A planar robotic lizard that uses four five-bar mechanisms (only RRRRR) for the head, body and tail is shown in our previous work \cite{rajashekhar2022design}. But this robot uses the uneven forward pushing force that is generated from the five-bar mechanisms to move only in the forward direction. This limitation is overcome in this work.      

Robotic mechanisms are classified as open chain \cite{wright2012design} and close chain \cite{chen2003locomotion} based on the arrangement of linkages. In some cases they can be a combination of two \cite{gao2015kinematics}. In order for a mechanism to function properly, its singularity positions have to be identified. A method to identify the configuration space singularities have been proposed in \cite{bandyopadhyay2004analysis}. It helps to derive the conditions at which the mechanism reaches the singularity conditions. Similar works have been reported in \cite{gosselin1990singularity,choudhury2000singularity}. The dimensions of the links in the mechanisms play an important role in determining the singularities. In this work we perform the dimensional synthesis and identify the singularity positions for the body and leg of the robotic lizard in order to avoid them during the functioning of the robot.  
    
\begin{figure}[h!]
\begin{center}
\includegraphics[scale=0.35]{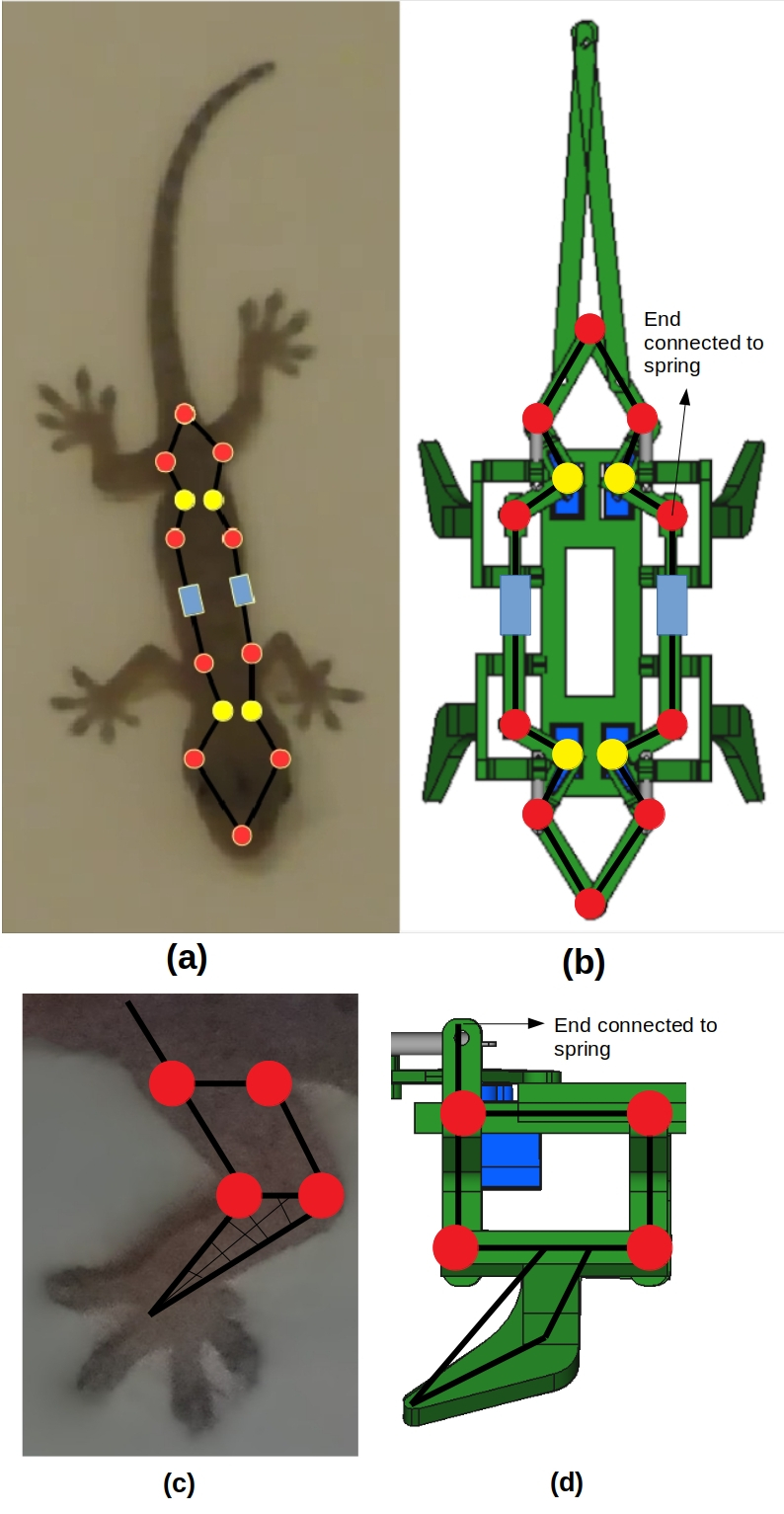}
\end{center}
\caption{The yellow and red circles denote the active and passive revolute joints respectively. The blue rectangle denotes the passive prismatic joint. (a) The common house gecko is chosen as an inspiration for bio-mimicking (b) The computer aided design of the body (c) The leg of a common house gecko taken as an inspiration (d) The computer aided design of the leg.}
\label{fig_intro}
\end{figure} 

\begin{figure}[h!]
\begin{center}
\includegraphics[scale=0.35]{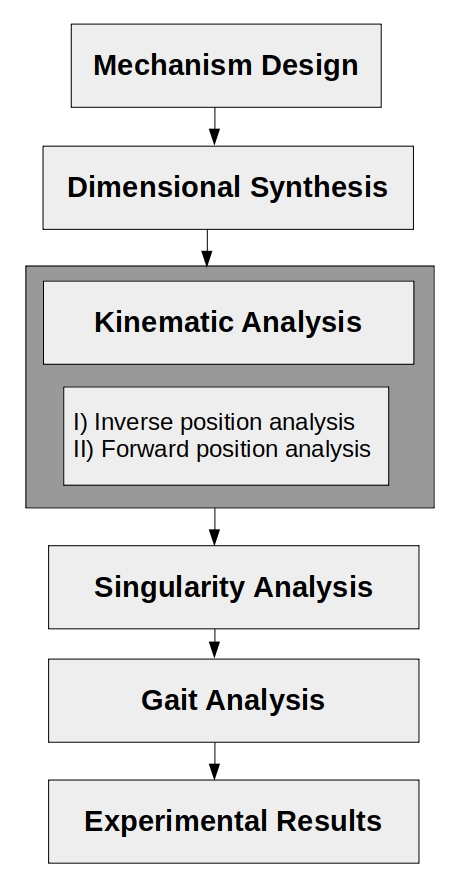}
\end{center}
\caption{The flowchart represents the optimal kinematic design procedure followed in this work.}
\label{fig_optimal_design_flowchart}
\end{figure}

In this work, a mechanism is designed to mimic the motion of the common house gecko as shown in Figure \ref{fig_intro}. A comparison is made between the body of the biological lizard and the robotic lizard in Figure \ref{fig_intro} (a,b). Similarly the leg of the biological lizard is compared with the robotic lizard in Figure \ref{fig_intro} (c,d). The novelty about this mechanism is that, the head and tail are actuated using four rotary actuators which in turn actuates the four legs individually through a spring. The robotic lizard was built using a 3D printer and experiments were conducted using the same. By sweeping a combination of the four actuators about a certain angle, the robot exhibits various gaits. The optimal kinematic design procedure is presented in Figure \ref{fig_optimal_design_flowchart}. The mechanism design is shown in Section \ref{sec_mec_des}. The dimensional synthesis of the mechanism is done in Section \ref{sec_dim_syn} followed by the kinematic analysis in Section \ref{sec_fwd_kin}. Then the singularity identification of the mechanism is done in Section \ref{sec_sing_iden} to find the gain type singularities of the mechanism. Then the gait analysis is done within the specified range of angular motion of the actuators in Section \ref{sec_gait_ana}. The experimental results are presented in Section \ref{sec_experi_res}. The concluding remarks are given in Section \ref{sec_conc}. The video of the robotic lizard in action can be seen here: \hyperlink{https://youtu.be/FpZ50tB2Tm8}{https://youtu.be/FpZ50tB2Tm8}

\section{Mechanism Design}
\label{sec_mec_des}
\begin{figure}[h!]
\begin{center}
\includegraphics[scale=0.35]{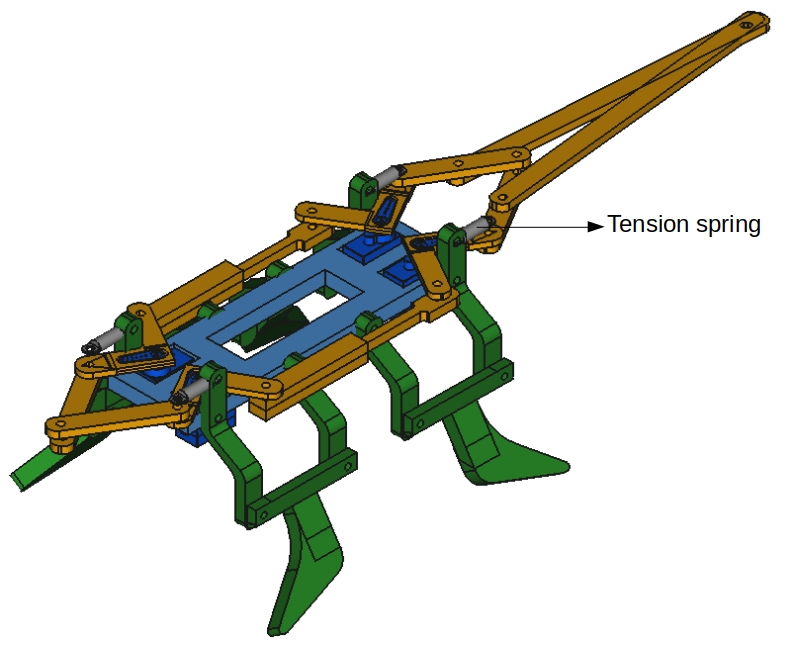}
\end{center}
\caption{The sub-system I and sub-system II are shown in orange and green shades respectively. The tension spring connects the sub-system I to II at four places. The chassis is shown in blue shade.}
\label{fig_subsystem}
\end{figure}

Linkage mechanisms have been used to mimic lizards in the past. For instance, in the works of \cite{dai2009biomimetics}, gecko-like robot has been made with rigid body linkages and rotary actuators. In this work, isometric view of the robotic lizard mechanism is shown in Figure \ref{fig_subsystem}. The mechanism consists of two sub-systems. One is the body mechanism (sub-system I) which is shown in the figure in orange shade and the other is the leg mechanism (sub-system II) shown in figure in green shade. These two sub-systems are connected through tension springs which transfers motion from sub-system I to II. The rotary actuators are placed on the chassis which is shown in blue shade.

\begin{figure}[h!]
\begin{center}
\includegraphics[scale=0.5]{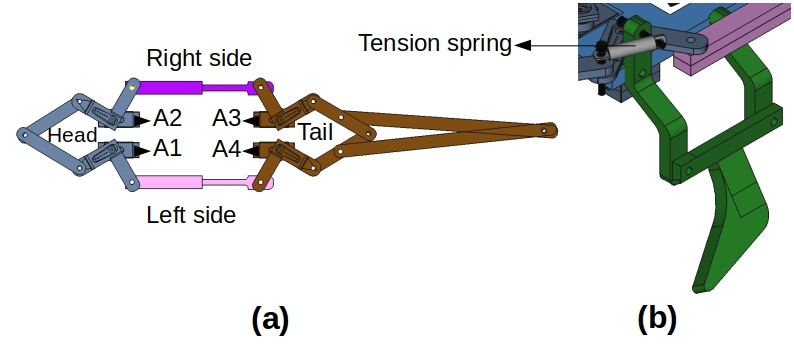}
\end{center}
\caption{(a) The sub-system I: The head, left body, right body and tail are shown in grey, pink, magenta and brown shades respectively (b) The sub-system II: The four-bar mechanism is shown is green shade. The tension spring transfers the motion from the sub-system I to II.}
\label{fig_mech_design}
\end{figure} 

\subsection{Sub-system I: Body Mechanism}
The body mechanism is made up of four five-bar mechanisms as shown in Figure \ref{fig_mech_design} (a). It is made up of 13 links of which 8 are binary, 4 are ternary and 1 is quarternary in nature. There are 14 revolute and 2 prismatic joints that connect these links to form the body mechanism. The long tail links at the end are neglected as they do not contribute to the degrees of freedom. The degrees of freedom (M) is calculated as follows,
\begin{equation}
\label{equ_dof}
M=3(N-1-j)+\sum_{i=1}^{j}f_{i}
\end{equation}   
where,\\
$N$=Number of links\\
$j$=Number of joints\\
$\sum_{i=1}^{j}f_{i}$=Sum of degrees of freedom of each joint\\ 
In our case, substituting the known values, it is found that the sub-system has 4 degrees of freedom. The four active revolute joints are shown as yellow circles in Figure \ref{fig_intro} (a). The other passive joints in the tail including 3 binary joints and 2 prismatic joints (in the body) are shown as red circles and blue rectangles respectively. 

The active links are the four ternary links which are connected to the actuators (A1, A2, A3 and A4) as shown in Figure \ref{fig_mech_design} (a). When the active joints (A1 and A2) in the head five-bar mechanism (RRRRR) are actuated, the head moves accordingly. This in-turn transfers the motion to the sub-system II through the tension springs. In the same way, when the active joints (A3 and A4) in the tail five-bar mechanism (RRRRR) are actuated, the tail moves accordingly. This also transfers the motion to the sub-system II through the tension springs. The left body (RRPRR) five-bar mechanism is actuated by active joints A1 and A4. Similarly the right body (RRPRR) five-bar mechanism is actuated by active joints A2 and A3.

\subsection{Sub-system II: Leg Mechanism}
The leg mechanism is consists of four four-bar mechanisms as shown in Figure \ref{fig_subsystem} in green shade. Each four-bar mechanism is made up of four links of which all are binary in nature. There are four binary joints that connect these links to form the mechanism. The fixed link of each four-bar mechanism is attached to the chassis. The degrees of freedom (M) is calculated as from Equation \ref{equ_dof} to be one. Therefore one tension spring can be used for each four-bar mechanism as a source of actuation. One end of the spring is connected to the ternary link of sub-system I and the other end is connected to the binary link in sub-system II as shown in Figure \ref{fig_mech_design} (b). As the active links of the sub-system I moves, the leg mechanism also moves in such a way that it goes front and back. This makes the robot to walk. There are two reasons why tension springs are needed in the mechanism. When the actuator A1 in the sub-system I sweeps, it does not transfer continuous motion to the sub-system II. The distance between the ternary link (active link in the sub-system I) and the link in four-bar mechanism (in sub-system II) varies as the actuators sweep. Due to these two reasons, the link that connects the two sub-systems is chosen to be a tension spring in order to maintain contact during motion.   

\section{Determining the Dimensions of the Linkages}
\label{sec_dim_syn}
 A path to be traced can be drawn in advance and the link lengths can be found based on it. It can be also done using neural networks \cite{vasiliu2001dimensional} and optimization techniques like Genetic Algorithms (GA) and Sequential Quadratic Programming (SQP) \cite{liu2018ga}. In this work, the dimensional synthesis is done as mentioned in \cite{liu2014parallel}. It is a Performance Chart-Based Design Methodology (PCbDM). For this the performance charts for the theoretical workspace, maximal inscribed circle and local conditioning index are plotted and presented in this section. 
\subsection{Performance Chart}
\begin{figure}[h!]
\begin{center}
\includegraphics[scale=0.6]{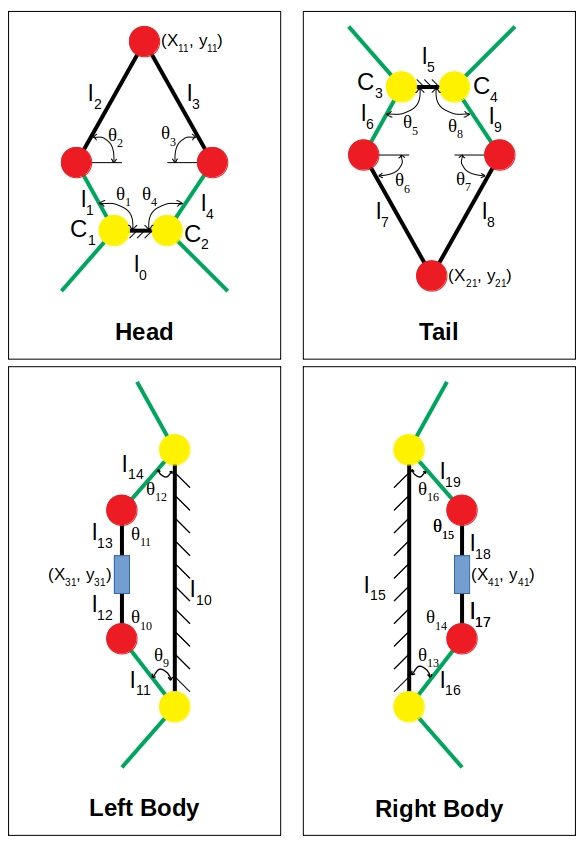}
\end{center}
\caption{Kinematic description of the robotic lizard body (Sub-System I)}
\label{fig_schematic_four_pics}
\end{figure}

\begin{figure}[h!]
\begin{center}
\includegraphics[scale=0.5]{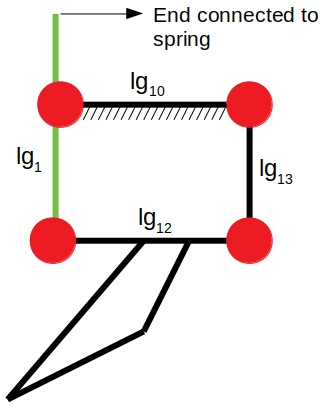}
\end{center}
\caption{Kinematic description of the robotic lizard leg (Sub-System II)}
\label{fig_schematic_leg}
\end{figure}

The kinematic description of the robotic lizard body is shown in Figure \ref{fig_schematic_four_pics}. The various parameters used for the dimensional synthesis of the body are presented in it. Similarly the various parameters of the leg mechanism is presented in Figure \ref{fig_schematic_leg}. The link in the green shade denote the active link in the mechanism. The constrains posed by the links are presented in Section \ref{sec_fwd_pos_ana}.

\subsubsection{Parameter Design Space}
In Figure \ref{fig_schematic_four_pics} head, let $R_1=l_1$, $R_2=l_2$ and $R_3=l_0/2$. The five-bar mechanism is symmetrical in nature. Therefore only three parameters are needed for dimensional synthesis. The difficulty here is that these parameters can take values between zero to infinity. In our case we measure the values of non-dimensional parameters $r_i$ from the head of the biological lizard. The values are as follows:
\begin{equation}
\label{equ_r_1}
r_1=0.3
\end{equation}
\begin{equation}
\label{equ_r_2}
r_2=0.5
\end{equation}
\begin{equation}
\label{equ_r_3}
r_3=0.1
\end{equation}
Eliminating the physical link size of the mechanism, the dimensional factor (D) is calculated as follows:
\begin{equation}
D=\frac{R_1+R_2+R_3}{0.9}
\end{equation}
The non-dimensional parameters $r_i$ can now be written as 
\begin{equation}
\label{equ_dim_final}
r_1 = \frac{R_1}{D}, r_2 = \frac{R_2}{D}, r_3 = \frac{R_3}{D}
\end{equation}
This equation yields
\begin{equation}
\label{equ_r1_r2_r3}
r_1 + r_2 + r_3 = 0.9
\end{equation}
In equation \ref{equ_r1_r2_r3}, the non-dimensional parameters $r_i$ can theoretically
take any value between $0$ and $0.9$. The studies on workspace and singularity of five-bar mechanisms prove that $r_1$ and $r_2$ cannot be $0$. Also the value of $r_1 + r_2$ should not be less than $r_3$, or else the mechanism assembly would fail. Thus the parameters are,
\begin{equation}
0 < r_1, r_2 <0.9
\end{equation}  
\begin{equation}
0 < r_3 <0.45
\end{equation} 
Thus the non-dimensional parameters $r_i$ chosen in Equations \ref{equ_r_1}-\ref{equ_r_3} fall within this criteria. Therefore we proceed with these conditions. 

\subsubsection{Theoretical Workspace Area}
The workspace is the area where the end effector of the mechanism can reach to perform the intended task. In this case, the theoretical workspace of the head five-bar mechanism is governed by four circles. It is shown in Figure \ref{fig_lci_mic}. The region shown in pink shade is the theoretical reachable workspace. In our case only the upper region is going to be utilized and hence the LCI and MIC is calculated only for that region. The equations that form the four circles \cite{liu2014parallel} are as follows:
\begin{equation}
\left(x_{11} + l_{0}/2\right)^2 + y_{11} = \left(l_{1} + l_{2}\right)^2
\end{equation}
\begin{equation}
\left(x_{11} + l_{0}/2\right)^2 + y_{11} = \left(l_{1} - l_{2}\right)^2
\end{equation}
\begin{equation}
\left(x_{11} - l_{0}/2\right)^2 + y_{11} = \left(l_{1} + l_{2}\right)^2
\end{equation}
\begin{equation}
\left(x_{11} - l_{0}/2\right)^2 + y_{11} = \left(l_{1} - l_{2}\right)^2
\end{equation}
Similarly the workspace can be plotted for the tail, body and leg.

\subsubsection{Maximal Inscribed Circle}
It is shown in \cite{liu2014parallel} that the thickest part of the workspace is always symmetrical about the Y-axis. From the workspace diagram, we can observe that the circle satisfies this condition. The Maximal Inscribed Circle (MIC) inside the upper portion of the workspace is calculated as follows: 
\begin{equation}
x_{11}^2 + \left(y_{11} - y_{MIC}\right)^2 = r_{MIC}^2
\end{equation} 
where,
\begin{equation}
r_{MIC} = \left(l_1 + l_2 - |l_1 - l_2| \right)/2
\end{equation}
\begin{equation}
y_{MIC}=\sqrt{\left(l_1+l_2+|l_1-l_2|\right)^2 / 4 - \left(l_0/2\right)^2}
\end{equation}
This is plotted in the Figure \ref{fig_lci_mic} as a red circle. The value of $r_{MIC}=0.3$. This value plays a critical role in determining the boundaries for finding LCI.

\subsubsection{Local Conditioning Index}
\begin{figure}[h!]
\begin{center}
\includegraphics[scale=0.5]{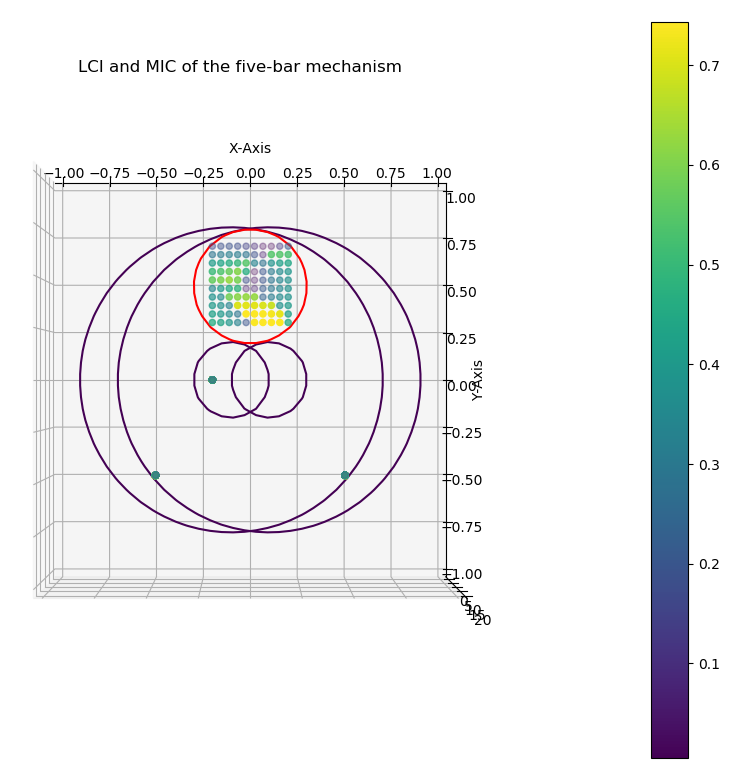}
\end{center}
\caption{The Local Conditioning Index (LCI) is shown inside the red circle. The Maximal Inscribed Circle (MIC) in the upper region of the five-bar mechanism is shown in red circle. The theoretical workspace is shown in pink shade.}
\label{fig_lci_mic}
\end{figure}

The Jacobian matrix is used to analyse the velocity and singularity conditions of the mechanisms. The accuracy with which the mechanism is controlled is determined by the condition number. The condition number of the Jacobian matrix is obtained from $\kappa =||J^{-1}||\hspace{2mm}||J||$. Here $||.||$ is the Euclidean norm of a matrix \cite{liu2014parallel}. The Local Conditioning Index (LCI) is defined as the inverse of the condition number. It is,
\begin{equation}
LCI = \frac{1}{\kappa}, 0 \le LCI \le 1
\end{equation}   
The value of LCI should be maintained as large as possible. The LCI helps in evaluation of static stiffness and dexterity of the five-bar mechanisms. The LCI of the head five-bar mechanism (using the non-dimensional parameters $r_i$) for our robotic lizard is shown in Figure \ref{fig_lci_mic}. It has a good LCI ranging more than $0.7$. This shows that the values that were chosen as the non-dimensional parameters $r_i$ is convincing. It is to be noted that the LCI was calculated for the head by considering the body and the tail simultaneously.

\subsection{Dimensional Synthesis}
\label{subsec_dim_sec}
The objective is to determine the dimensional factor $D$, using which the non-dimensional values $r_i$ can be converted to dimensional geometric parameters $l_i$. In our case we choose the $R_{3}$ as the parameter to find the value of \textit{D}. It is theoretically known that the head (end-effector) of the robotic lizard will operate only inside $R_{MIC}$ circle. Considering the dimension of $R_3 = 10mm$, since the minimum distance between the two servo motor (actuator A1 and A2) shafts $(l_{0}=20mm)$ is a constrain, we obtain from Equation \ref{equ_dim_final}, 
\begin{equation}
D = \frac{10}{0.1} =100
\end{equation}
By substituting the value of $D$ in Equation \ref{equ_dim_final}, we obtain the following:
\begin{equation}
R_1 = 30mm
\end{equation}
\begin{equation}
R_2 = 50mm
\end{equation}
The angle $C_i$ in the head and tail as shown in Figure \ref{fig_schematic_four_pics} is taken to be $90^\circ$ for better visualization of the robotic lizard. Thus we have successfully found the dimensions of the head of the robotic lizard. We can similarly find the dimensions of the tail, left and right body, and legs. For the leg of the robot, we have chosen parallelogram linkage (four-bar mechanism) since it does not operate in the singularity zone. Similar dimensional synthesis procedure was followed to obtain the dimensions of the leg mechanism. The dimensions are listed in Table \ref{tab_dim_syn}.

\begin{table}[h!]
\caption{The link lengths obtained after dimensional synthesis}
\label{tab_dim_syn}
\begin{tabular}{|c|c|}
\hline
\textbf{Link Name}                                                 & \textbf{\begin{tabular}[c]{@{}c@{}}Link Dimension\\ (mm)\end{tabular}} \\ \hline
l$_0$, l$_5$                                                       & 20                                                                     \\ \hline
l$_1$, l$_4$, l$_6$, l$_9$, l$_{14}$, l$_{11}$, l$_{16}$, l$_{19}$ & 30                                                                     \\ \hline
l$_2$, l$_3$, l$_7$, l$_8$                                         & 50                                                                     \\ \hline
l$_{12}$, l$_{13}$, l$_{16}$, l$_{18}$                             & 45                                                                     \\ \hline
l$_{10}$, l$_{15}$                                                 & 135                                                                    \\ \hline
lg$_{10}$, lg$_{12}$,                                              & 50                                                                     \\ \hline
lg$_{13}$, lg$_{1}$                                                & 45                                                                     \\ \hline
\end{tabular}
\end{table}

\section{Kinematic Analysis of the Robotic Lizard}
\label{sec_fwd_kin}
In this section we present the inverse and forward position analysis of the entire mechanism. We use geometric method for finding the inverse kinematics and vector loop method for finding the forward kinematics. The way how the sub-system I is kinematically connected with sub-system II is presented in the form of an algorithm. The velocity analysis and acceleration analysis are the next steps in the kinematic analysis. The velocity analysis is presented in Section \ref{sec_sing_iden} where it is combined with singularity identification. The acceleration analysis can be done differentiating the velocity equations with respect to time. This analysis is not presented in this work for the sake of brevity.     
\subsection{Inverse Position Analysis}
The inverse position analysis is done when the position and orientation of the output point is known and the input joint variables are unknown. In a five bar mechanism, two legs connect the output point and the base link. The rotary actuator is located in the base of the mechanism. We focus only on the inverse position analysis in this section because it aids in doing the Jacobian analysis. There are eight legs present in the parallel mechanisms that is formed by the body. Their constraint equations are written as follows.

Head:
\begin{equation}
\left(x_{11} - l_{1} \times \cos \theta_{1} + \left(\frac{l_{0}}{2}\right)\right)^2 + \left(y_{11} - l_{1} \times \sin \theta_{1}\right)^2=\left(l_{2}\right)^2
\end{equation}  
\begin{equation}
\left(x_{11} - l_{1} \times \cos \theta_{2} - \left(\frac{l_{0}}{2}\right)\right)^2 + \left(y_{11} - l_{1} \times \sin \theta_{2}\right)^2=\left(l_{2}\right)^2
\end{equation} 

Tail:
\begin{equation}
\left(x_{21} - l_{9} \times \cos \theta_{8} + \left(\frac{l_{5}}{2}\right)\right)^2 + \left(y_{21} - l_{9} \times \sin \theta_{8}\right)^2=\left(l_{8}\right)^2
\end{equation}  
\begin{equation}
\left(x_{21} - l_{9} \times \cos \theta_{7}  - \left(\frac{l_{5}}{2}\right)\right)^2 + \left(y_{21} - l_{9} \times \sin \theta_{7}\right)^2=\left(l_{8}\right)^2
\end{equation}

Left Body:
\begin{equation}
2 \times x_{31} = \left(l_{13} \times \cos \theta_{11} + l_{14} \times \cos \theta_{12}\right)
\end{equation}
\begin{equation}
2 \times y_{31} = \left(l_{13} \times \sin \theta_{11} + l_{14} \times \sin \theta_{12}\right)
\end{equation}

Right Body:
\begin{equation}
2 \times x_{41} = \left(l_{19} \times \cos \theta_{16} + l_{18} \times \cos \theta_{15}\right)
\end{equation}
\begin{equation}
2 \times y_{41} = \left(l_{19} \times \sin \theta_{16} + l_{18} \times \sin \theta_{15}\right)
\end{equation}

The sub-system II is made up of four-bar mechanisms. The inverse kinematics procedure for this mechanism is well known and is readily available in \cite{norton2009kinematics}. Using this one can easily obtain the inverse kinematic solution for the mechanism. Hence it is not presented here in this work.  

\subsection{Forward Position Analysis}
\label{sec_fwd_pos_ana}
Kinematic analysis for a gecko has been done in \cite{nam2009kinematic} by considering a combination of closed and open serial mechanisms. Similarly in our work, we combine the mechanisms in sub-system I with sub-system II and present their forward position analysis. The Figure \ref{fig_schematic_four_pics} and \ref{fig_schematic_leg} shows the various parameters of the two sub-systems. As mentioned earlier, we use the vector loop method to do the forward position analysis. The vector loop for the head is presented as follows by referring to Figure \ref{fig_schematic_four_pics} head:

\begin{equation}
\label{equ:head}
R_{1}+R_{2}-R_{3}-R_{4}-R_{0}=0
\end{equation}
This can be written in the scalar form as Equations \ref{equ:cos} and \ref{equ:sin}. 
\begin{equation}
\label{equ:cos}
l_{1}\cos(\theta_{1})+l_{2}\cos(\theta_{2})-l_{3}\cos(\theta_{3})-l_{4}\cos(\theta_{4})-l_{0}=0 \\
\end{equation}
\begin{equation}
\label{equ:sin}
l_{1}\sin(\theta_{1})+l_{2}\sin(\theta_{2})-l_{3}\sin(\theta_{3})-l_{4}\sin(\theta_{4})=0
\end{equation}
The $\theta_{2}$ and $\theta_{3}$ are obtained by rearranging the equations \ref{equ:cos} and \ref{equ:sin}.
\begin{equation}
\label{equ:theta4}
\theta_3=2\arctan\frac{-B-\sqrt{B^{2}-4AC}}{2A}
\end{equation}
\begin{equation}
\label{equ:theta3}
\theta_2=2\arctan\frac{-E+\sqrt{E^{2}+4DF}}{2D}
\end{equation}
where,\\
$K_1=l_{4}\sin(\theta_{4})-l_{1}\sin(\theta_{1})$;\\
$K_2=l_{4}\cos(\theta_{4})-l_{1}\cos(\theta_{1})+l_{0}$\\
$K_3=-l_{4}\sin(\theta_{4})+l_{1}\sin(\theta_{1})$;\\
$K_4=-l_{4}\cos(\theta_{4})+l_{1}\cos(\theta_{1})+l_{0}$\\
$A=\frac{{l_{3}}^{2}-2\,K_2\,l3-{l_{2}}^{2}+{K_2}^{2}+{K_1}^{2}}{2}$;\\
$B=2\,K_1\,l_{3}$;\\
$C=\frac{{l_{3}}^{2}+2\,K_2\,l_{3}-{l_{2}}^{2}+{K_2}^{2}+{K_1}^{2}}{2}$\\
$D=\frac{{l_{3}}^{2}-{l_{2}}^{2}+2\,K_4\,l_{2}-{K_4}^{2}-{K_3}^{2}}{2}$;\\
$E=2\,K_3\,l_{2}$;\\
$F=\frac{{l_{3}}^{2}-{l_{2}}^{2}-2\,K_4\,l_{2}-{K_4}^{2}-{K_3}^{2}}{2}\\$

Similarly, the other vector loop equations for the rest of the body can be formed as follows by changing the orientation of the parts accordingly.\\

For tail: 
\begin{equation}
\label{equ:tail}
R_{6}+R_{7}-R_{8}-R_{9}-R_{5}=0
\end{equation}
For left side of the body:
\begin{equation}
\label{equ:leftbody}
R_{11}+R_{12}-R_{13}-R_{14}-R_{10}=0
\end{equation}
For right side of the body:
\begin{equation}
\label{equ:rightbody}
R_{16}+R_{17}-R_{18}-R_{19}-R_{15}=0
\end{equation}

There are links that are common in head, left and right body, and tail five-bar mechanisms. They can be represented as constrain equations. It is as follows:
\begin{equation}
\theta_{10}+\theta_{11}=0
\end{equation}
\begin{equation}
\theta_{14}+\theta_{15}=0
\end{equation}
\begin{equation}
\theta_{1}-\theta_{12}+C_{1}=0
\end{equation}
\begin{equation}
\theta_{4}-\theta_{16}+C_{2}=0
\end{equation}
\begin{equation}
\theta_{9}-\theta_{5}+C_{3}=0
\end{equation}
\begin{equation}
\theta_{13}-\theta_{8}+C_{4}=0
\end{equation}

It is to be noted that the links $l_{0}$, $l_{5}$, $l_{10}$, $l_{15}$, $lg_{10}$, $lg_{20}$, $lg_{30}$ and $lg_{40}$ lay on the chassis of the robot.

Referring to Figure \ref{fig_schematic_leg}, the vector loop equations for the leg can be written as,
\begin{equation}
\label{equ:leg1}
LG_{1}+LG_{12}-LG_{13}-LG_{10}=0
\end{equation}

The $\theta_{lg12}$ and $\theta_{lg13}$ are obtained by rearranging the scalar equations that are obtained from Equation \ref{equ:leg1} which is similar to \ref{equ:cos} and \ref{equ:sin}.

\begin{equation}
\label{equ_lg_12}
\theta_{lg12}=2\arctan\left(\frac{-K-\sqrt{K^{2}-4JL}}{2J}\right)
\end{equation}

\begin{equation}
\label{equ_lg_13}
\theta_{lg13}=2\arctan\left(\frac{-H+\sqrt{H^{2}-4GI}}{2G}\right)
\end{equation}

where,\\
$G=\cos(\theta_{lg_1})-\frac{lg_{10}}{lg_1}-\frac{lg_{10}\cos(\theta_{lg1})}{lg_{13}}+\frac{{lg_1}^2+{lg_{13}}^2+lg_{10}^2-lg_{12}^2}{2lg_{1}lg_{13}} $\\
$H=-2\sin(\theta_1)$\\
$I=\frac{lg_{10}}{lg_{1}}-(\frac{lg_{10}}{lg_{13}}+1)\cos(\theta_{lg1})+\frac{lg_{1}^2+lg_{13}^2+lg_{10}^2-lg_{12}^2}{2lg_{1}lg_{13}}$\\
$J=\cos(\theta_{lg1})-\frac{lg_{10}}{lg_{1}}-\frac{lg_{10}\cos(\theta_{lg1})}{lg_{12}}+\frac{lg_{13}^2-lg_{10}^2-lg_{1}^2-lg_{12}^2}{2lg_{1}lg_{12}}$\\
$K=-2\sin(\theta_{lg1})$\\
$L=\frac{lg_{10}}{lg_{1}}-(\frac{lg_{10}}{lg_{12}}+1)\cos(\theta_{lg1})+\frac{lg_{13}^2-lg_{10}^2-lg_{1}^2-lg_{12}^2}{2lg_{1}lg_{12}}\\$

The vector loop equations for the other legs can be written as follows by changing the orientation of the legs accordingly.

\begin{equation}
\label{equ:leg2}
LG_{2}+LG_{22}-LG_{23}-LG_{10}=0
\end{equation}
\begin{equation}
\label{equ:leg3}
LG_{3}+LG_{32}-LG_{33}-LG_{10}=0
\end{equation}
\begin{equation}
\label{equ:leg4}
LG_{4}+LG_{42}-LG_{43}-LG_{10}=0
\end{equation}

Thus we have derived the forward and inverse position equations for the two sub-sections. The way in which they are connected are explained in the following section.
\subsection{Connecting the two sub-systems}
In this section we show using an algorithm how the two sub-systems are connected. This is useful for forward position analysis and gait analysis of the robotic lizard. It is presented in Algorithm \ref{algo_fwd_kin}. The $\theta_1$, $\theta_4$, $\theta_5$ and $\theta_8$ are driven by actuators $A1$, $A2$, $A4$ and $A3$ respectively. Using these known (given) input angles, the unknown angles are calculated from the equations in Section \ref{sec_fwd_pos_ana}.

\begin{algorithm}
\caption{The Kinematic connection between the two sub-systems}\label{algo_fwd_kin}
\begin{algorithmic}
\For {$\theta_1$, $\theta_4$ = given input} \Comment{Head five-bar mechanism}
\State find $\theta_2$ and $\theta_3$ from Equations \ref{equ:theta4} and \ref{equ:theta3}
\For {$\theta_1$ = $\theta_{lg1}$} \Comment{Front left leg four-bar mechanism}
\State find $\theta_{lg12}$ and $\theta_{lg13}$ from Equations \ref{equ_lg_12} and \ref{equ_lg_13}
\For {$\theta_4$ = $\theta_{lg2}$} \Comment{Front right leg four-bar mechanism}
\State find $\theta_{lg22}$ and $\theta_{lg23}$
\For {$\theta_5$, $\theta_8$ = given input} \Comment{Tail five-bar mechanism}
\State find $\theta_6$ and $\theta_7$
\For {$\theta_5$ = $\theta_{lg4}$} \Comment{Back left leg four-bar mechanism}
\State find $\theta_{lg42}$ and $\theta_{lg43}$
\For {$\theta_8$ = $\theta_{lg3}$} \Comment{Back right leg four-bar mechanism}
\State find $\theta_{lg32}$ and $\theta_{lg33}$
\For {$\theta_{12}$ = $C_1$ + $\theta_1$ and $\theta_9$ = $\theta_5$ - $C_3$} \Comment{Left body}
\State find $\theta_{10}$
\State assign $\theta_{10}$ = $-\theta_{11}$
\For {$\theta_{16}$ = $C_2$ + $\theta_4$ and $\theta_{13}$ = $\theta_8$ - $C_4$} \Comment{Right body}
\State find $\theta_{14}$
\State assign $\theta_{14}$ = $-\theta_{15}$
\EndFor
\EndFor
\EndFor
\EndFor
\EndFor
\EndFor
\EndFor
\EndFor
\end{algorithmic}
\end{algorithm}

\section{Singularity Identification}
\label{sec_sing_iden}
The loop closure equations were differentiated with respect to the respective joint variables and then seperated into two matrices $[K]$ and $[K^*]$. The $[K]$ matix contains the active joint variables and the $[K^*]$ matrix consists of the passive joint variables. This method was adopted from \cite{bandyopadhyay2004analysis}. Equating the determinant of $[K]$ and $[K^*]$ matrix to $0$, we obtain the singularity condition.
\subsection{Singularity condition for the Sub-system I}
\begin{figure}[h!]
\begin{center}
\includegraphics[scale=0.60]{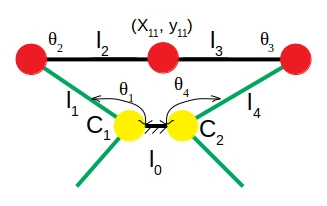}
\end{center}
\caption{The singular configuration of the five-bar mechanism when $\sin\left(\theta _{2}+\theta _{3}\right) = 0$. The dimensions of the links should be chosen in such a way that this configuration does not occur or the input angles should be carefully given to avoid this configuration.}
\label{fig_singu_fbm}
\end{figure}

The $[K]$ matrix which contains the geometric details of active joint variables is as follows: 

    $ [K]=  \left(\scalemath{0.7}{\begin{array}{cccccccc} -l_{1}\,\sin\left(\theta _{1}\right) & -l_{4}\,\sin\left(\theta _{4}\right) & 0 & 0 & 0 & 0 & 0 & 0\\ l_{1}\,\cos\left(\theta _{1}\right) & -l_{4}\,\cos\left(\theta _{4}\right) & 0 & 0 & 0 & 0 & 0 & 0\\ 0 & 0 & -l_{6}\,\sin\left(\theta _{5}\right) & -l_{9}\,\sin\left(\theta _{8}\right) & 0 & 0 & 0 & 0\\ 0 & 0 & l_{6}\,\cos\left(\theta _{5}\right) & -l_{9}\,\cos\left(\theta _{8}\right) & 0 & 0 & 0 & 0\\ 0 & 0 & 0 & 0 & l_{11}\,\sin\left(C_{3}-\theta _{5}\right) & -l_{14}\,\sin\left(C_{1}+\theta _{1}\right) & 0 & 0\\ 0 & 0 & 0 & 0 & l_{11}\,\cos\left(C_{3}-\theta _{5}\right) & -l_{14}\,\cos\left(C_{1}+\theta _{1}\right) & 0 & 0\\ 0 & 0 & 0 & 0 & 0 & 0 & l_{16}\,\sin\left(C_{4}-\theta _{8}\right) & -l_{19}\,\sin\left(C_{2}+\theta _{4}\right)\\ 0 & 0 & 0 & 0 & 0 & 0 & l_{16}\,\cos\left(C_{4}-\theta _{8}\right) & -l_{19}\,\cos\left(C_{2}+\theta _{4}\right) \end{array}}\right)$
The determinant of $[K]$ can be ensured to be a value other than zero or infinite by carefully controlling the input values to the actuators. The $[K^*]$ matrix which contains the geometric details of passive joint variables is as follows:
 
    $[K^*]=\left(\scalemath{0.7}{\begin{array}{cccccccc} -l_{2}\,\sin\left(\theta _{2}\right) & -l_{3}\,\sin\left(\theta _{3}\right) & 0 & 0 & 0 & 0 & 0 & 0\\ l_{2}\,\cos\left(\theta _{2}\right) & -l_{3}\,\cos\left(\theta _{3}\right) & 0 & 0 & 0 & 0 & 0 & 0\\ 0 & 0 & -l_{7}\,\sin\left(\theta _{6}\right) & -l_{8}\,\sin\left(\theta _{7}\right) & 0 & 0 & 0 & 0\\ 0 & 0 & l_{7}\,\cos\left(\theta _{6}\right) & -l_{8}\,\cos\left(\theta _{7}\right) & 0 & 0 & 0 & 0\\ 0 & 0 & 0 & 0 & -l_{13}\,\sin\left(\theta _{11}\right) & l_{12}\,\sin\left(\theta _{11}\right) & 0 & 0\\ 0 & 0 & 0 & 0 & l_{13}\,\cos\left(\theta _{11}\right) & l_{12}\,\cos\left(\theta _{11}\right) & 0 & 0\\ 0 & 0 & 0 & 0 & 0 & 0 & -l_{17}\,\sin\left(\theta _{14}\right) & l_{18}\,\sin\left(\theta _{14}\right)\\ 0 & 0 & 0 & 0 & 0 & 0 & l_{17}\,\cos\left(\theta _{14}\right) & l_{18}\,\cos\left(\theta _{14}\right) \end{array}}\right)$
Taking the determinant and equating it to $0$ gives the singularity condition. It is as follows.
\begin{equation}
\label{equ_singu_five_bar}
    \frac{l_{2}\,l_{3}\,l_{7}\,l_{8}\,l_{12}\,l_{13}\,l_{17}\,l_{18}\,\sin\left(2\,\theta _{11}\right)\,\sin\left(2\,\theta _{14}\right)\,\left(\cos\left(\theta _{2}+\theta _{3}-\theta _{6}-\theta _{7}\right)-\cos\left(\theta _{2}+\theta _{3}+\theta _{6}+\theta _{7}\right)\right)}{2}=0
\end{equation}  
    The significance of this equation is that, when $\sin\left(2\,\theta _{11}\right)$ or $\sin\left(2\,\theta _{14}\right)$ or $\sin\left(\theta _{2}+\theta _{3}\right)$ or $\sin\left(\theta _{6}+\theta _{7}\right)$ is $0$ or when a combination of them is $0$, then the mechanism is said to be in singular configuration. It should also be noted that the link lengths of the links mentioned in the above Equation \ref{equ_singu_five_bar} should have a finite value to avoid the singularity condition. The singularity condition is shown in Figure \ref{fig_singu_fbm}.

\subsection{Singularity condition for the Sub-system II}
The $[K]$ and $[K^*]$ matrices of the sub-system II are as follows.

$  [K]=  \left(\begin{array}{cc} -lg_{12}\,\sin\left(\theta _{lg1}+\theta _{lg12}\right) & -lg_{1}\,\sin\left(\theta _{lg1}\right)\\ lg_{12}\,\cos\left(\theta _{lg1}+\theta _{lg12}\right) &  lg_{1}\,\cos\left(\theta _{lg1}\right) \end{array}\right)
$

$  [K^*]=  \left(\begin{array}{cc} lg_{13}\,\sin\left(\theta _{lg13}\right) & -lg_{12}\,\sin\left(\theta _{lg1}+\theta _{lg12}\right)\\ -lg_{13}\,\cos\left(\theta _{lg13}\right) & lg_{12}\,\cos\left(\theta _{lg1}+\theta _{lg12}\right) \end{array}\right)$

Taking the determinant of $[K^*]$ and equating it to $0$ gives the singularity condition. It is as follows.

    $-lg_{12}\,lg_{13}\,\sin\left(\theta _{lg1}+\theta _{lg12}-\theta _{lg13}\right)=0$
    
The significance of this equations is that when $\left(\theta _{lg1}+\theta _{lg12}-\theta _{lg13}\right)$= $0^\circ$ or $180^\circ$ the mechanism is structurally singular.  

\section{Gait Representation}
\label{sec_gait_ana}
It is observed in the works of \cite{farley1997mechanics} that the mechanics of locomotion of lizard and other legged animals are similar in nature. The study also reveals that lizards use two gaits, namely mammalian walking and trotting. Gait planning using kinematics was done in \cite{son2010gait}. Motion capture technique was used to analyze the trot gait in lizard \cite{kim2013trotting}. The trot gait parameters of the robotic lizard has been obtained by simulation in \cite{kim2014trot}. In this work the gait analysis was done by setting the joint range for the input angles in the Algorithm \ref{algo_fwd_kin}.

\subsection{Walking Gait}
Walking gaits have been studied using reinforcement algorithms \cite{mock2023comparison} and PID control \cite{alnasrallah2023statically}. In this paper only preliminary works have been done for performing the walking gait by the robotic lizard. It is based on the work done in \cite{barron2010hardware}.
    
\subsection{Trotting Gait}
Quadrupeds with springy legs that exhibit trotting gait have been done in the past \cite{li2014control}. The Central Pattern Generators (CPG) were used to generate the trot gait in \cite{zhang2014trot}. As mentioned in the previous section, the trotting gait is also based on the work done in \cite{barron2010hardware}.

\section{Experimental Results}
\label{sec_experi_res}
In this section we present the design and experimental results that arise out of the robotic lizard. We first explain how the robot is designed and fabricated. Then we discuss about the experimental set-up followed by the gaits exhibited by the robot. 
\subsection{Design and Fabrication of the Robot}
\begin{figure}[h!]
\begin{center}
\includegraphics[scale=0.7]{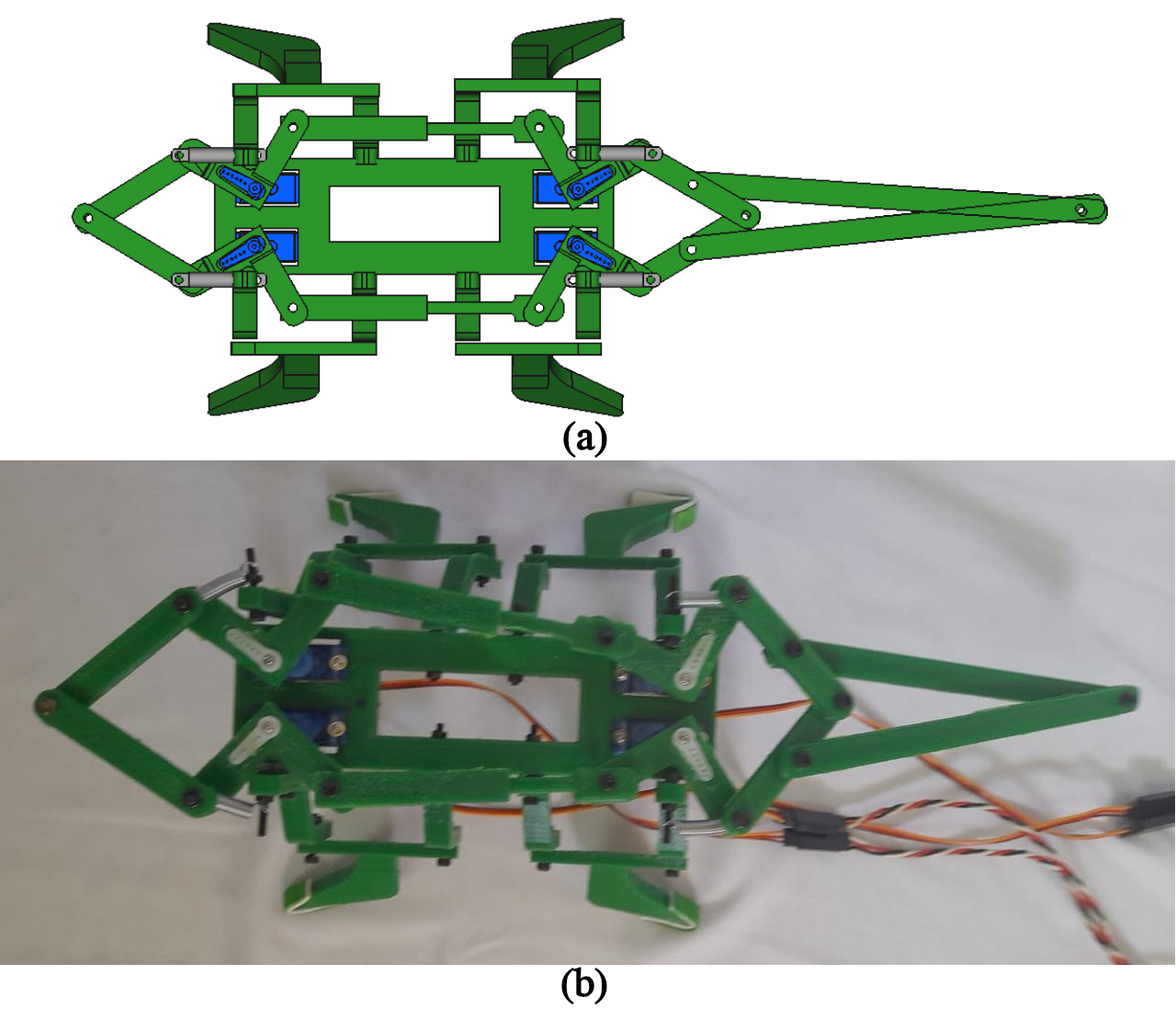}
\end{center}
\caption{(a) The CAD of the robotic lizard (b) The fabricated robotic lizard}
\label{fig_cad_real}
\end{figure}

The Computer-Aided-Design (CAD) of the robot was done using \textit{FreeCAD$^{TM}$} software. The lengths of the links were based on the dimensions that were obtained in Section \ref{subsec_dim_sec}. The parts were assembled to form the robotic lizard. It is shown in Figure \ref{fig_cad_real} (a). These parts were printed using \textit{Creality Ender 3} 3D printer. The parts were assembled to form the prototype which is shown in Figure \ref{fig_cad_real} (b).

\subsection{Experimental Setup}
The chassis was designed and is shown in blue shade in Figure \ref{fig_subsystem}. The four servo motors \textit{(SG-90 servo motors)} are mounted on the chassis. The ternary links are connected to these servo motors and from there the other links are connected. The holes that accommodate the joints are of standard size \textit{M3}. There are two long links that are attached to tail of the body in order to make it appear close to the biological lizard. The leg of the robot (tip that touches the ground) is stuck with a rubber pad for better contact with the floor while walking. The two sub-systems are connected using a tension spring which has the specification: \textit{Ext (5.55 $\times$ 25.4)}. 
      
\subsection{Gaits Exhibited by the Robotic Lizard}
\begin{figure}[h!]
\begin{center}
\includegraphics[scale=0.35]{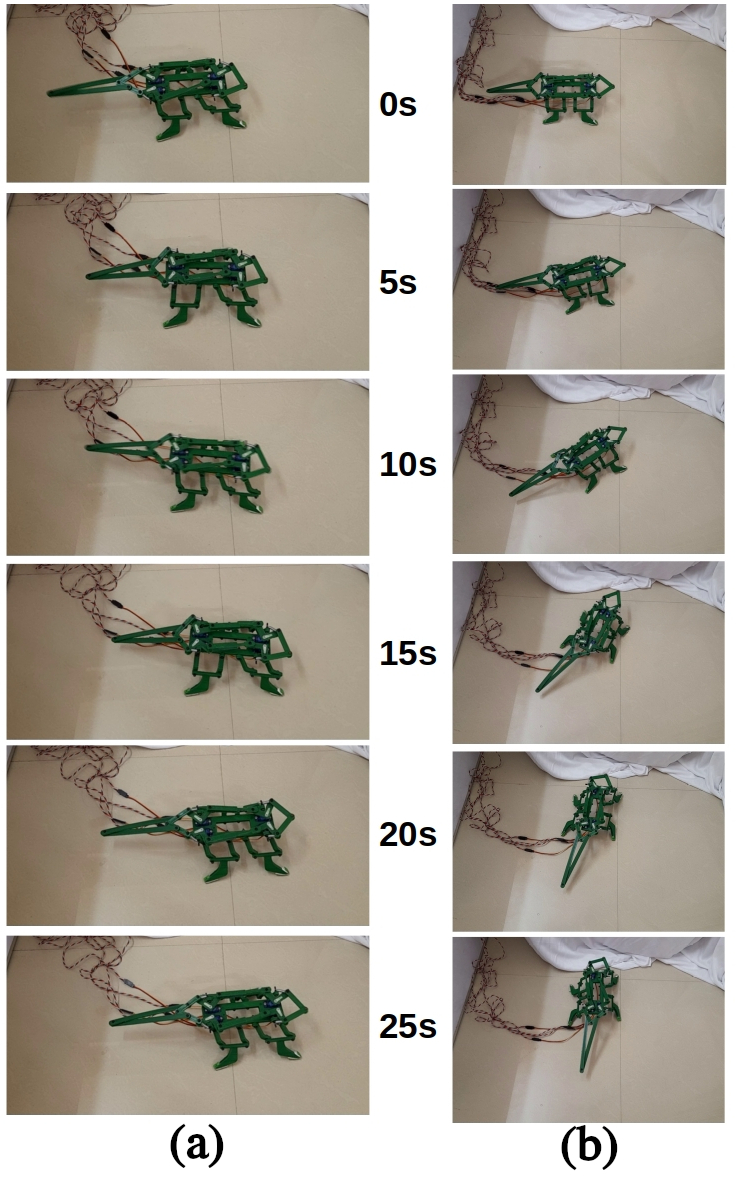}
\end{center}
\caption{The gaits exhibited by the robotic lizard (a) Walking forward (b) Turing towards left}
\label{fig_experi_gaits}
\end{figure}

The robotic lizard is made to exhibit walking and trotting gaits. It is also made to turn left about an axis. The \textit{Arduino UNO} is used as a micro-controller to control the robotic lizard. The program was written for these gaits based on the data from \cite{barron2010hardware}. The walking gait is shown in Figure \ref{fig_experi_gaits} (a) for about \textit{25s}. The left turning motion is done and the results are shown in \ref{fig_experi_gaits} (b). It is observed that the robot walks slowly when it moves on a straight line than while taking a turn. This could be due to the imbalance in the contact forces between the foot and the floor.      

\section{Conclusions}
\label{sec_conc}
In this work, we presented a robotic lizard that mimics a common house Gecko. The mechanism was optimally designed by considering the robot as two sub-systems. In this work we have done dimensional synthesis, kinematic analysis and singularity identification of the robotic lizard mechanism that was designed newly by us. To the best of the authors knowledge, the robotic lizard designed in this work closely resembles the biological lizards in their morphology. In the future other performance parameters like Good Condition Workspace (GCW) will be considered for the mechanism. Although the lizards are known for climbing walls and ceiling, in this work we limit it to walking on flat surfaces. In future we will attempt to make it climb walls and ceilings.

\bibliography{sn-bibliography}

\end{document}